\documentclass[letterpaper, 10 pt, conference]{ieeeconf}  % Comment this line out if you need a4paper

\IEEEoverridecommandlockouts                              % This command is only needed if 
\usepackage{graphicx} % for pdf, bitmapped graphics files
\usepackage[dvipsnames]{xcolor}

\title{\LARGE \bf
A Review on Objective-Driven Artificial Intelligence
}

\author{Apoorv Singh$^{1}$ % <-this % stops a space
\thanks{$^{1}$Apoorv Singh is from Robotics Institute, Carnegie Mellon University (CMU)
{\tt\small apoorv93singh@gmail.com}}%
}
\begin{document}
\maketitle
\thispagestyle{empty}
\pagestyle{empty}

%%%%%%%%%%%%%%%%%%%%%%%%%%%%%%%%%%%%%%%%%%%%%%%%%%%%%%%%%%%%%%%%%%%%%%%%%%%%%%%%
\begin{abstract}
While advancing rapidly, Artificial Intelligence still falls short of human intelligence in several key aspects due to inherent limitations in current AI technologies and our understanding of cognition. Humans have an innate ability to understand context, nuances, and subtle cues in communication, which allows us to comprehend jokes, sarcasm, and metaphors. Machines struggle to interpret such contextual information accurately. Humans possess a vast repository of common-sense knowledge that helps us make logical inferences and predictions about the world. Machines lack this innate understanding and often struggle with making sense of situations that humans find trivial. In this article, we review the prospective Machine Intelligence candidates, a review from Prof. Yann LeCun, and other work that can help close this gap between human and machine intelligence. Specifically, we talk about what's lacking with the current AI techniques such as supervised learning, reinforcement learning, self-supervised learning, etc. Then we show how Hierarchical planning-based approaches can help us close that gap and deep-dive into energy-based, latent-variable methods and Joint embedding predictive architecture methods. 

\end{abstract}

%%%%%%%%%%%%%%%%%%%%%%%%%%%%%%%%%%%%%%%%%%%%%%%%%%%%%%%%%%%%%%%%%%%%%%%%%%%%%%%%
\section{INTRODUCTION}
Closing the gap between machine and human intelligence is a complex and multidimensional challenge encompassing various fields, including artificial intelligence (AI), cognitive science, neuroscience, and philosophy. The goal is to create machines that emulate human-like cognitive abilities, decision-making, and problem-solving capabilities.

Machine learning, particularly deep learning, has shown remarkable success in solving specific tasks when provided with sufficient labeled data for training. These systems excel in tasks like image classification \cite{efficientnet}, language translation \cite{criss}, speech recognition \cite{speach}, and playing specific games \cite{alphago}.

However, these AI systems lack the broad and flexible intelligence that humans possess. They struggle with adapting to new and unforeseen scenarios, understanding context, reasoning with common sense, transferring knowledge from one domain to another, and possessing a comprehensive understanding of the world. Moreover, these AI systems require training on huge datasets at extreme computation costs. However, humans can learn new tasks very quickly with their common sense. They understand how to world works and can reason and plan accordingly. 

\textbf{Problems with the current Machine Learning approaches:} In this section we go through each major sub-divisions of the ML approach and discuss their major fundamental issues.
\begin{itemize}
    \item Supervised Learning: It relies heavily on labeled data for training, and if the data is insufficient, unrepresentative, or of poor quality, it can lead to poor model performance. This can result in overfitting, where the model performs well on the training data but poorly on unseen data. Moreover, this form of ML only leads us to a task-specific model, which lacks generalizability to understand the world's knowledge.
    \item Reinforcement Learning: This requires extreme amounts of trials, and it still depends on one of the encoded rewards and policies that we set. Due to this, ensuring that an RL agent can generalize its learned policy to new, unseen environments or situations is complex. Training in one environment might not lead to good performance in a slightly different environment.
    \item Generative AI: Generative predictions only work well for text and other discrete modalities. Currently, it is impossible to generate perfect images, let alone videos, or even solve autonomous vehicles.
\end{itemize}

\begin{figure}[ht]
\vskip -0.1in
\begin{center}
\centerline{\includegraphics[width=\columnwidth]{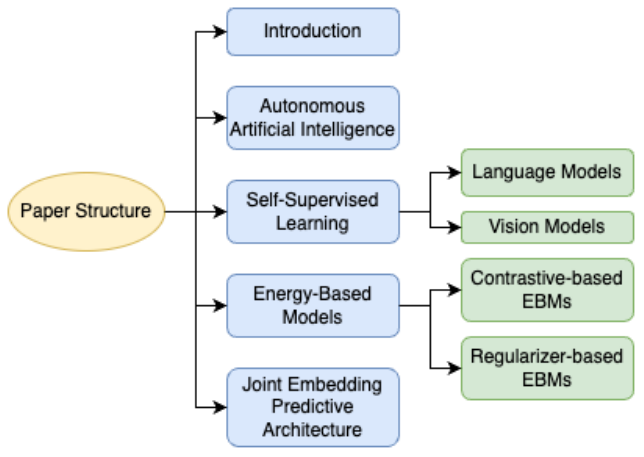}}
\caption{Structure of the Survey paper}
\label{fig:structure}
\end{center}
\vskip -0.2in
\end{figure}

To build autonomous AI systems, we must ask ourselves: \emph{How could machines learn like animals and humans?}. Humans can generalize their learning from one domain to another with relative ease. AI models often struggle to transfer knowledge learned in one task to another, requiring extensive retraining for each new task. As shown in Fig. \ref{fig:structure}, in the rest of the paper, we will cover broad latest research areas that Machine Learning researchers are developing to close this gap.

\section{Autonomous Artificial Intelligence}
Autonomous artificial intelligence is an advancement in current AI technology, although its AI cannot develop an awareness of its environment or itself as humans do. But autonomous AI can make its own decisions in various emergencies or is highly critical. Autonomous AI orchestrates the different technologies needed to solve a problem in any of these cases and uses the most relevant ones at its discretion, based on parameters set by humans. \\ \\ 
There are two main modes of operation for an Autonomous AI system. Firstly, \emph{Perception-Action Cycle}. This mode of operation first extracts the representation of the state of the work through perception. Then it goes to the policy module and is transformed into the final action. You can call it more of a reactive system where you perceive the surroundings and then act upon them. The other mode of operation is \emph{Perception-Planning-Action Cycle}. This mode of operation is more analytical, as used also on MPC (Model Predictive Control). The added planning module is where the actor proposes multiple actions to take in addition to the cost associated with each of them. Based on these costs, hierarchical planning is performed, discussed in the later section, then action is performed.

\section{Self-Supervised Learning}
One of the everyday things that human learning and machine learning have in common is self-supervised learning. Self-supervised learning has shown its application on multiple fronts, from Language models \cite{bert} to computer vision models \cite{mae, dino} and protein folding models \cite{protein}. 

The basic idea behind self-supervised learning is to design tasks that require the model to understand relationships, context, or features within the data. These tasks generate surrogate labels or supervisory signals from the data, allowing the model to learn meaningful representations without relying on explicit human-generated labels. These learned representations, a.k.a foundational models, can then be used as a foundation for solving downstream tasks, often with less labeled data. However, It's important to note that while self-supervised learning reduces the need for labeled data, it still requires careful design of the self-supervised tasks to ensure that the learned representations are meaningful and generalizable.
\subsection{Language Models}
Bidirectional Encoder Representation from Transformers (BERT) \cite{bert} from the Google AI team has become a gold standard when it comes to several NLP tasks such as natural language inference, question answering, and more. It is a multi-task network that has an encoder-only model. It  uses 12 and 24 transformer layers in the Base and Large model, respectively. The way that BERT works is that it is trained on two different tasks. The first task is called the masked language model, where sentences are masked (typically around 15\%), and the model is trained to predict masked words. The right balance of this masking \% is essential as too little masking will make the training process extremely masking, and too much masking removes the context as required. The second task is to predict the following sentence. For example, the model is given two sets of sentences, and BERT learns the relationship between the sentences and predicts the next sentence given the first one. BERT is responsible if sentence B is the following sentence given sentence A. This is a binary classification task. This helps BERT to perform at the sentence level. To train BERT, three types of embeddings are required: token, segment, and position embeddings. Token embedding represents each word in a vector representation which is a fixed-size embedding. Segment embedding holds semantic information used to make sense of the semantic information for binary classification of the sentences, as seen above. Position embedding is to learn the order of the sentence by representing the order in a fixed-size positional embedding space. BERT is designed to process input sequences up to a length of 512. BERT can be used for single-sentence classification, sentence pair classification, question answering, and single-sentence tagging tasks. 

\subsection{For vision models}
\begin{figure}[ht]
\vskip -0.1in
\begin{center}
\centerline{\includegraphics[width=\columnwidth]{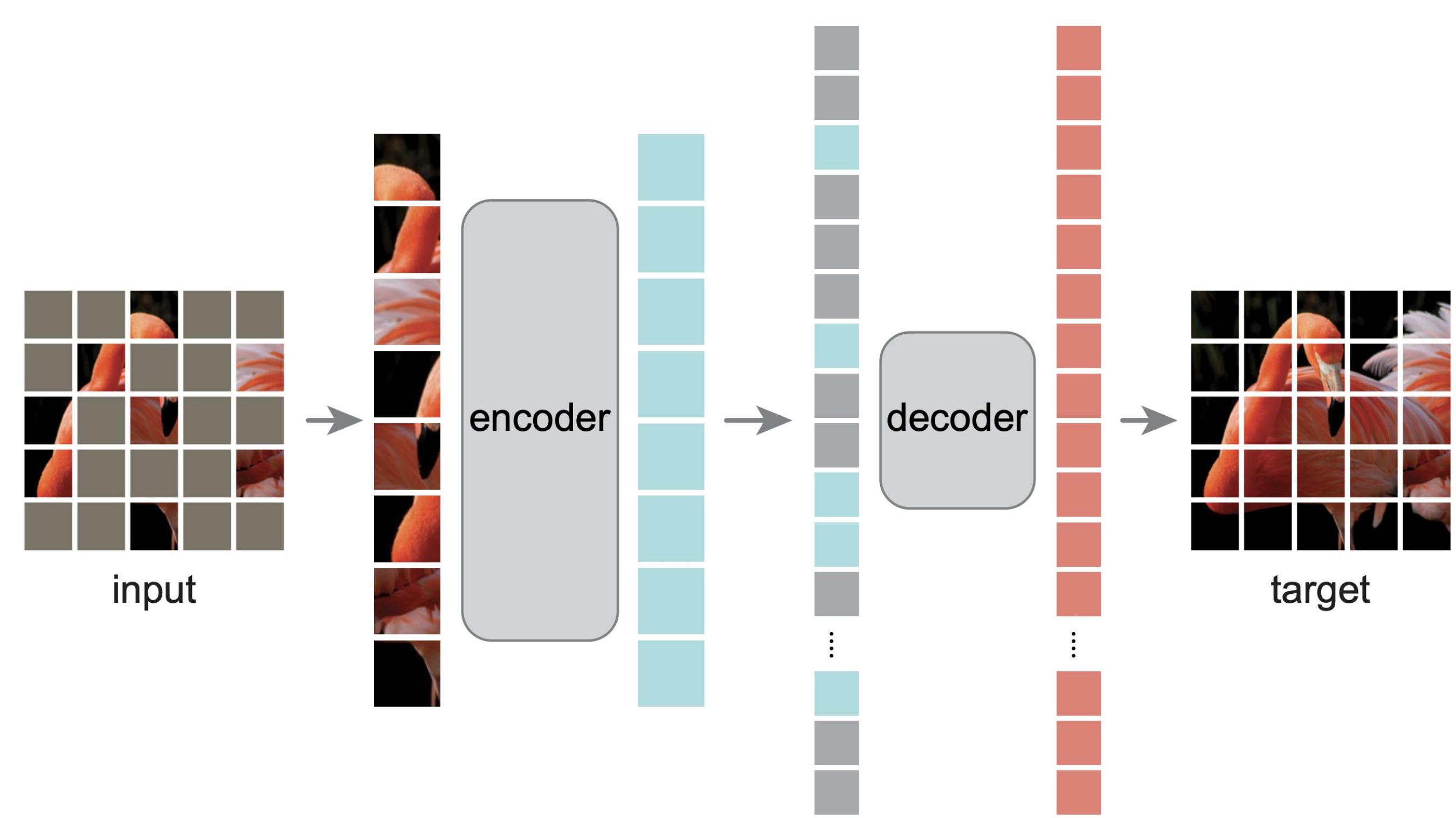}}
\caption{Visualization to show how self-supervised learning is learned using hidden tokens in MAE \cite{mae}.}
\label{fig:mae}
\end{center}
\vskip -0.2in
\end{figure}

 One typical example can be seen with Masked Autoencoders \cite{mae}, where they developed an asymmetric encoder-decoder architecture. A large random subset of image patches (e.g., 75\%) is masked during pre-training. The encoder is applied to the small subset of visible patches and a lightweight decoder that reconstructs the original image from the latent representation and mask tokens. After pre-training, the decoder is discarded, and the encoder is applied to uncorrupted images (full sets of patches) for the recognition task, as shown in Fig. \ref{fig:mae}. Here, authors, simple random patches are masked instead of attempting to remove objects. These patches most likely do not form semantic segments. However, one extension of this approach could be to apply an instance segmentation task \cite{maskrcnn} over the image and mask object-specific pixels, which later the decoder is trained to predict.

\section{Energy-Based Models}
\begin{figure}[ht]
\vskip -0.1in
\begin{center}
\centerline{\includegraphics[width=150pt]{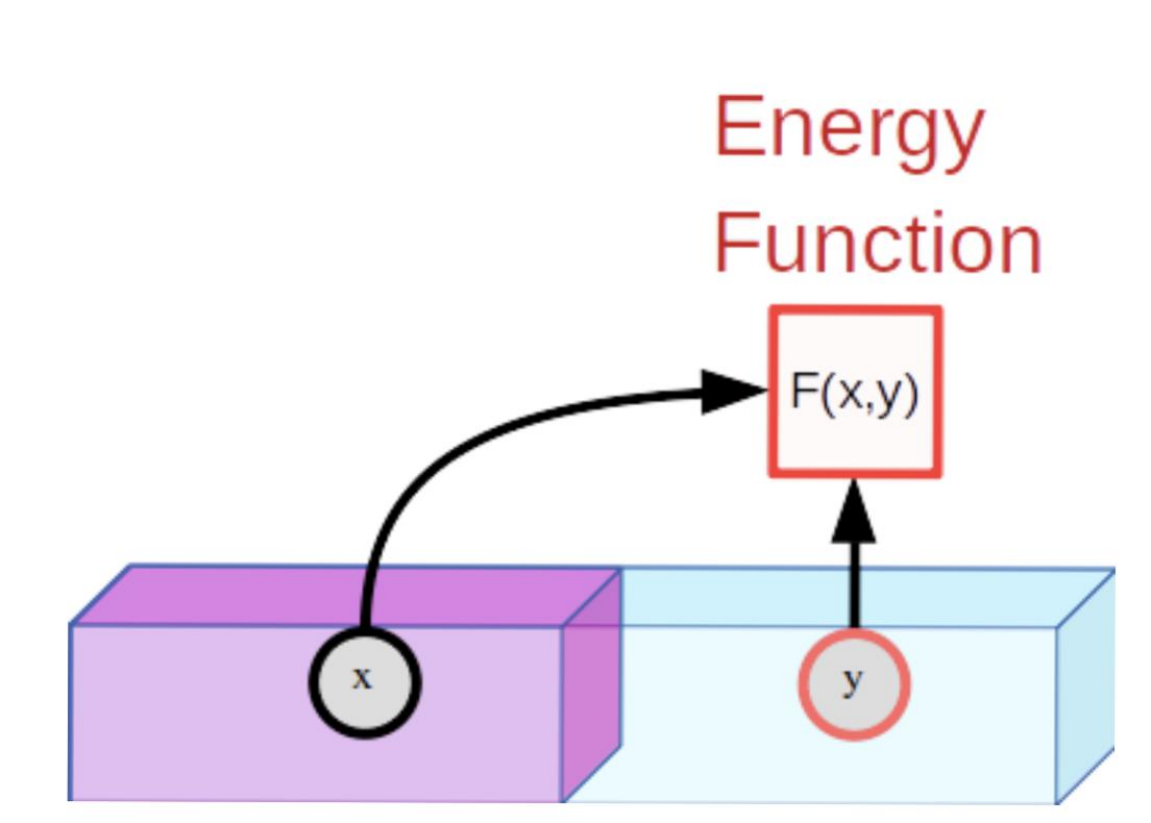}}
\caption{Simple demonstration for an EBM model, which calculates energy (similarity) between two inputs (purple and blue).}
\label{fig:ebm}
\end{center}
\vskip -0.2in
\end{figure}

Energy-based models (EBM) are one way to look at self-supervised models. Given two inputs, $x$ and $y$, EBMs can be used to tell their compatibility with energy between them, where low energy is associated with compatibility and vice versa, as shown in Fig. \ref{fig:ebm}. This $x$ and $y$ pair could be words, sentences, or even images. Similar to contrastive learning \cite{constrastive}, two things need to be taken care of in an EBM:
\begin{itemize}
    \item Train on compatible pairs of $x$ and $y$ to produce low energy between them.
    \item Ensure that incompatible pairs of $x$ and $y$ produce high energy. This is a more difficult and hence more interesting part.
\end{itemize}

Failure to perform the second task above is referred to as the collapse of the model. Two ways to avoid these collapses: 
\subsection{Contrastive energy-based SSL}
The core idea behind a contrastive model is to encourage the model to minimize the distance between similar examples while maximizing the distance between dissimilar examples in the learned representation space. This is typically achieved through a loss function that quantifies the difference between pairs of examples. One of the common ways to implement a contrastive model is by using Siamese networks or triplet networks:
\begin{itemize}
    \item \textbf{Siamese Networks:} Siamese networks consist of two identical subnetworks (often called "arms"), which share weights and parameters. Each arm takes in one input example and produces a representation. The representations from both arms are then compared using a distance metric (e.g., Euclidean distance or cosine similarity). The network is trained using a contrastive loss function that pushes similar examples to have lower energy and dissimilar examples to have higher energy in the scalar representation space.
    \item \textbf{Triplet Networks:} Triplet networks use sets of three examples: an anchor, a positive example (similar to the anchor), and a negative example (dissimilar to the anchor). The goal is to minimize the energy between the anchor and the positive example while maximizing the energy between the anchor and the negative example. The network is trained by optimizing a triplet loss function that captures these relationships.
\end{itemize}

These contrastive methods have a major issue: They are very inefficient to train. In high-dimensional spaces such as images, one image can be different from another in many ways. Finding a set of contrastive images covering how they can differ from a given image is nearly impossible.

\subsection{Regularizer-based SSL}
These methods minimize the volume of space that can take low energy. It consists of constructing a loss function that pushes down on the energies of training samples and simultaneously minimizes the volume of y space to which the model associates low energy. The volume of the low-energy region is measured by a regularization term in the energy/ loss. By minimizing this regularization term while pushing down on the energies of data points, the regions of low energy will “shrink-wrap” the regions of high data density. The main advantage of non-contrastive regularized methods is that they are less likely than contrastive methods to fall victim to the curse of dimensionality. \\
These approaches are more actively developed in the literature with BYOL \cite{byol}, DeeperCluster \cite{cluster}, and many more. BYOL \cite{byol} from DeepMind performs image representation with the SSL approach, where they have two neural networks, referred to as \textit{online} and \textit{target} networks, that interact and learn from each other. These two networks have the same architecture but different weights. From an augmented view of an image, the online network is trained to predict the target network representation of the same image under a different augmented view. At the same time, the target network is updated with a slow-moving average of the online network. BYOL does NOT need any negative samples since it is NOT trained using contrastive loss. In contrast to this approach, DeeperCluster \cite{cluster} proposes a new unsupervised approach that leverages self-supervision and clustering to capture complementary statistics from large-scale data.

There is an extension of this work called latent variable EBMs. In these models, energy depends on three variables: $x$ and $y$, as discussed earlier, and a latent variable $z$ which is not provided as an input. Intuitively this $z$ capturing of information is required to predict the compatibility of $x$ and $y$, which is not present in the two. For example, let's take $x$ as a view of a scene in an RGB image, and $y$ is the same scene but with a gray-scale image. Here $z$ can be the information that tells us to look at the scene by color-invariance scale to predict low energy between $x$ and $y$.  \\ \\ 
For the future, As iterated by Prof. Yann LeCun, "The challenge of the next few years may be to devise non-contrastive methods for a latent-variable energy-based model that successfully produce good representations of image, video, speech, and other signals and yield top performance in downstream supervised tasks without requiring large amounts of labeled data."

\section{JEPA: Joint Embedding Predictive Architecture}
To our knowledge, the JEPA term was first coined by Prof. Yann LuCun. This model takes in two inputs, $x$ and $y$, for which we are trying to capture dependency of. Then these inputs are passed in through separate encoders $Enc(x)$ and $Enc(y)$. We train the system so that the representation of $y$, i.e., $S_y$, is easily predictable from the representation of $x$, i.e., $S_x$. Then we have  a trainable predictor, which may or may not have a latent variable $z$, and the goal of it is to predict $S_y'$, given $S_x$ and $z$. \\
Note, here, we don't have a model that predicts $y'$ but a model that predicts $S_y'$. This is very important, as it allows our model to prune out all the irrelevant information in $y$ to make the right prediction. For example, in autonomous vehicle prediction, our task is to predict the on-road agents' (cars/ pedestrians) future. Here we only care about that specific information, and we don't care about whatever is happening outside the drivable surface, for example, a pedestrian on a balcony at a hotel. With our model, like JEPA, it can completely ignore the calculation of such future trajectories of the pedestrian at a hotel balcony and then reduce the complication of the problem by keeping it objective-driven. This architecture can be seen in VICReg \cite{vicreg}.

\begin{figure}[ht]
\vskip -0.1in
\begin{center}
\centerline{\includegraphics[width=\columnwidth]{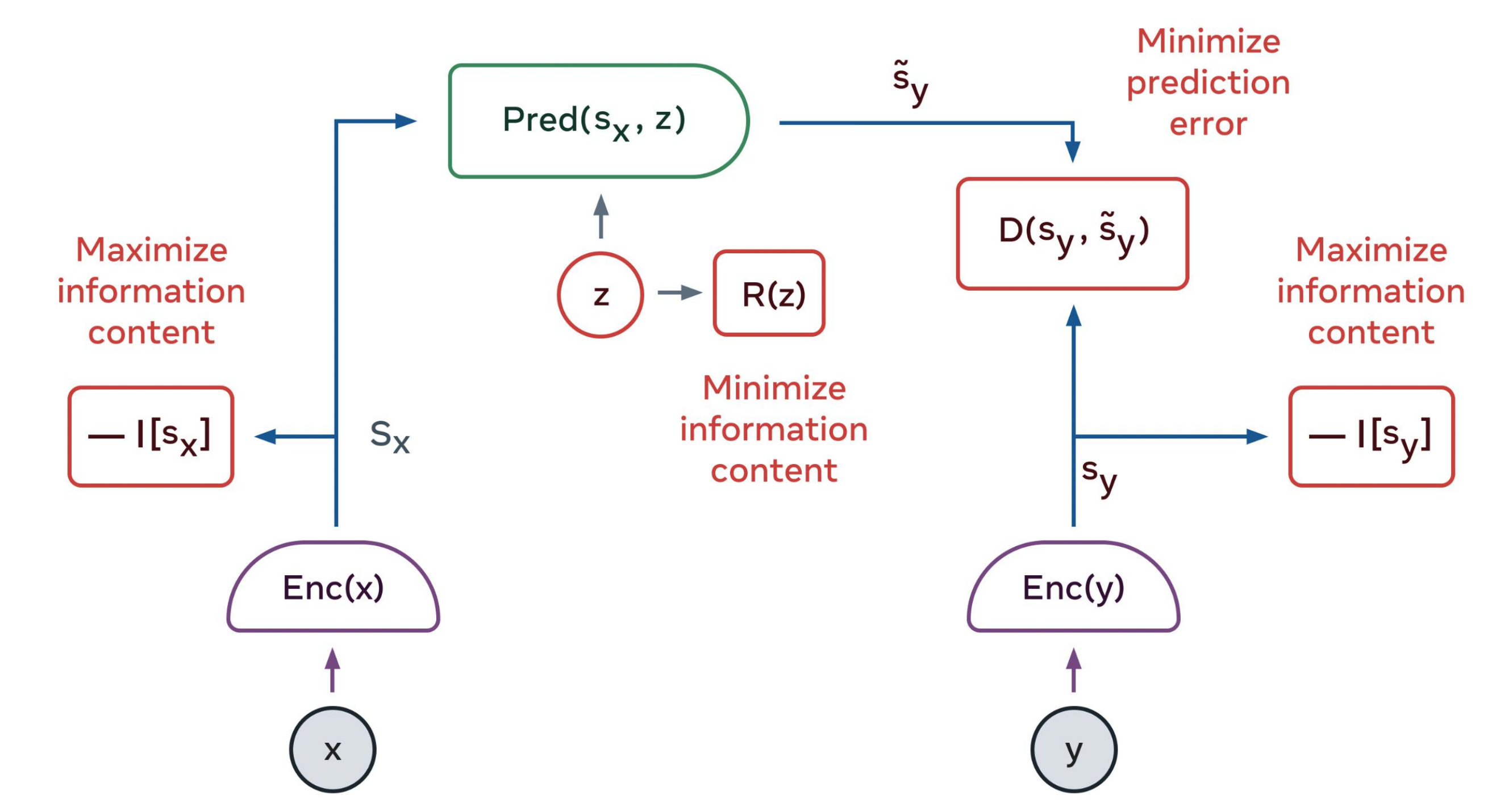}}
\caption{A simple representation of a JEPA model.}
\label{fig:jepa}
\end{center}
\vskip -0.2in
\end{figure}

To train this JEPA, we have four objective functions: 
\begin{itemize}
    \item $-I(S_x)$: Maximize information content in encoded representation, $S_x$ has for $x$.
    \item $-I(S_y)$: Maximize information content in encoded representation, $S_y$ has for $y$.
    \item $D (S_y, \tilde{S_y})$: Minimize the prediction error, such that the representation of y, $S_y$ is easily predictable from the representation of x, $S_x$. This is where the system is trained to leave out non-required information like pedestrians on the balcony.
    \item $R(z)$: Minimize the information in the latent variable, $z$, else $z$ can completely take over the training (without even looking at $x)$, and the model will collapse. This can be assumed as a regularizer.
    The latent variable contains everything you don't know about the world. This is information that we can not extract from the $x$. To find the value of $z$, we ask what amount of information is needed to complement $S_x$ to predict $S_y$.  
\end{itemize}

\section{Hierarchical Planning}
\begin{figure}[ht]
\vskip -0.1in
\begin{center}
\centerline{\includegraphics[width=\columnwidth]{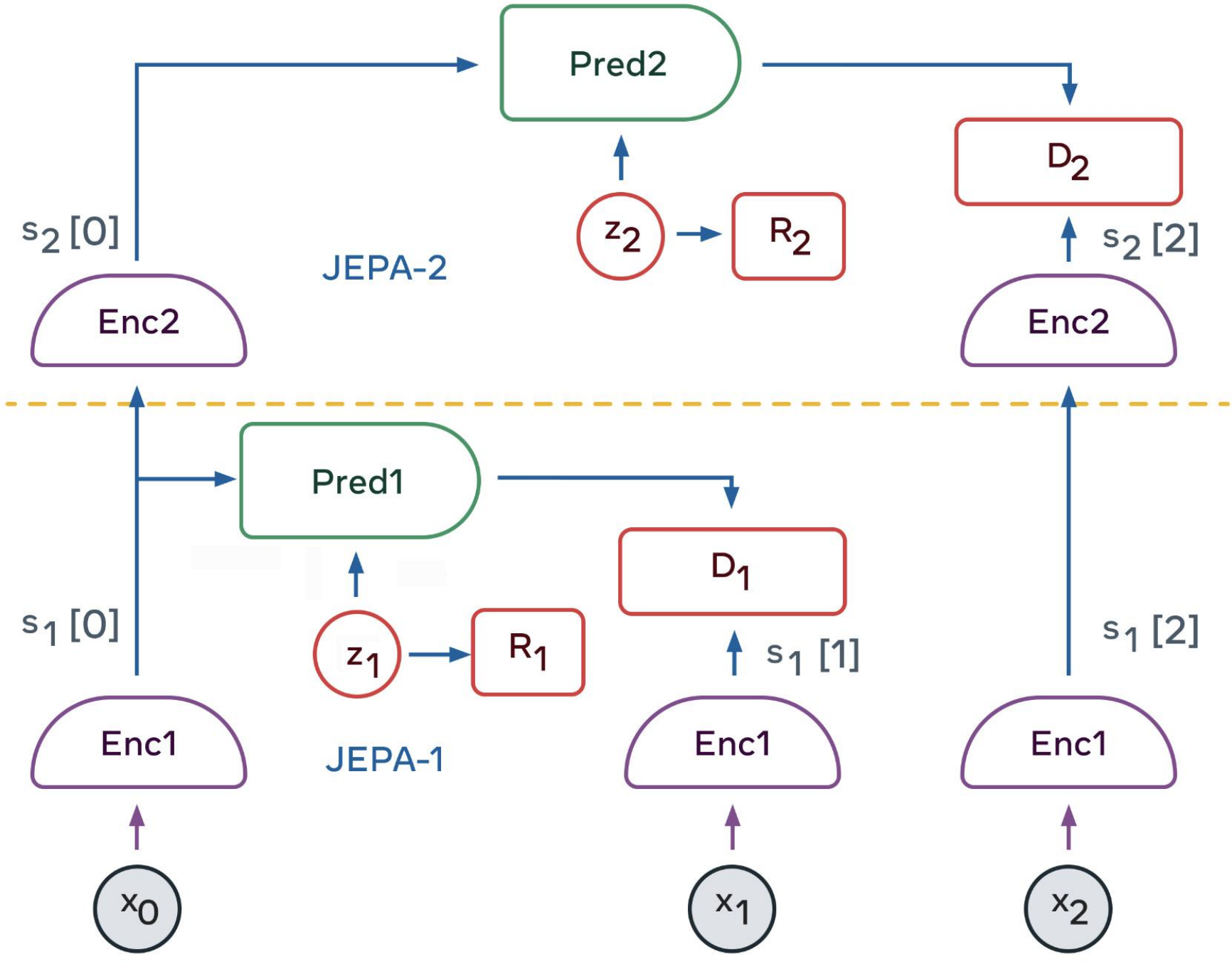}}
\caption{Heirarchial Jepa (two-stage)}
\label{fig:heirarchial}
\end{center}
\vskip -0.2in
\end{figure}

To solve the long-term prediction problem, this is a further extension of JEPA work. The main idea is for short-term prediction; we need a lot of low-level information about the scene, for example, trajectory prediction in autonomous driving. However, for long-term prediction, we need to abstract away more information to only look at the high-level information required for a holistic scene understanding. Here, the concept is stacking another layer of the JEPA on top of the other layer so that the second encoder can further encode information of $x$ to abstract the information further. This encoding of $x$ has even less information about $x$, but with that abstract but relevant information, we may be able to make even longer predictions. \\
To summarize: Low-level representation can only predict in the short term, as too many details make predictions hard. However, higher-level representation can predict in the longer term with fewer details, as shown in Fig. \ref{fig:heirarchial}. This concept can be extensively used in Hierarchical planning, where we have a big goal, but to achieve it, we need to decompose it into a series of short-term goals to get there.

As stated by many, this kind of hierarchical planning model is how humans operate and can lead to machine intelligence close the gap with human intelligence. There hasn't been much work around this topic yet, but it seems a very reasonable candidate to reason and plan a more complex task than a simple word prediction/ image classification.

\section{Conclusion}
In conclusion, as a first step, we have to get self-supervised learning to work to get models of the world with infinite data. Given these models are sophisticated enough, the system might be able to gather some common sense like humans/ animals. Architecture must be built on top of JEPA and Energy-based architectures, which follow optimal control principles, not generative AI or reinforcement learning. Also, we need to focus on hierarchical models, the founding principles of deep learning, which say how to selectively and incrementally abstract away information in the latent representation.

\addtolength{\textheight}{-12cm}   % This command serves to balance the column lengths

%%%%%%%%%%%%%%%%%%%%%%%%%%%%%%%%%%%%%%%%%%%%%%%%%%%%%%%%%%%%%%%%%%%%%%%%%%%%%%%%

%%%%%%%%%%%%%%%%%%%%%%%%%%%%%%%%%%%%%%%%%%%%%%%%%%%%%%%%%%%%%%%%%%%%%%%%%%%%%%%%

%%%%%%%%%%%%%%%%%%%%%%%%%%%%%%%%%%%%%%%%%%%%%%%%%%%%%%%%%%%%%%%%%%%%%%%%%%%%%%%%

{\small
\bibliographystyle{ieee_fullname}
\bibliography{IEEEexample}

\begin{thebibliography}{10}\itemsep=-1pt

\bibitem{vicreg}
Adrien Bardes, Jean Ponce, and Yann LeCun.
\newblock Vicreg: Variance-invariance-covariance regularization for
  self-supervised learning.
\newblock {\em arXiv preprint arXiv:2105.04906}, 2021.

\bibitem{cluster}
Mathilde Caron, Piotr Bojanowski, Julien Mairal, and Armand Joulin.
\newblock Unsupervised pre-training of image features on non-curated data.
\newblock In {\em Proceedings of the IEEE/CVF International Conference on
  Computer Vision (ICCV)}, October 2019.

\bibitem{dino}
Mathilde Caron, Hugo Touvron, Ishan Misra, Herv{\'{e}} J{\'{e}}gou, Julien
  Mairal, Piotr Bojanowski, and Armand Joulin.
\newblock Emerging properties in self-supervised vision transformers.
\newblock {\em CoRR}, abs/2104.14294, 2021.

\bibitem{constrastive}
Ting Chen, Simon Kornblith, Mohammad Norouzi, and Geoffrey Hinton.
\newblock A simple framework for contrastive learning of visual
  representations.
\newblock In {\em International conference on machine learning}, pages
  1597--1607. PMLR, 2020.

\bibitem{bert}
Jacob Devlin, Ming-Wei Chang, Kenton Lee, and Kristina Toutanova.
\newblock Bert: Pre-training of deep bidirectional transformers for language
  understanding.
\newblock {\em arXiv preprint arXiv:1810.04805}, 2018.

\bibitem{byol}
Jean-Bastien Grill, Florian Strub, Florent Altch{\'e}, Corentin Tallec, Pierre
  Richemond, Elena Buchatskaya, Carl Doersch, Bernardo Avila~Pires, Zhaohan
  Guo, Mohammad Gheshlaghi~Azar, et~al.
\newblock Bootstrap your own latent-a new approach to self-supervised learning.
\newblock {\em Advances in neural information processing systems},
  33:21271--21284, 2020.

\bibitem{mae}
Kaiming He, Xinlei Chen, Saining Xie, Yanghao Li, Piotr Doll{\'a}r, and Ross
  Girshick.
\newblock Masked autoencoders are scalable vision learners.
\newblock In {\em Proceedings of the IEEE/CVF conference on computer vision and
  pattern recognition}, pages 16000--16009, 2022.

\bibitem{maskrcnn}
Kaiming He, Georgia Gkioxari, Piotr Doll{\'a}r, and Ross Girshick.
\newblock Mask r-cnn.
\newblock In {\em Proceedings of the IEEE international conference on computer
  vision}, pages 2961--2969, 2017.

\bibitem{alphago}
David Silver, Aja Huang, Christopher Maddison, Arthur Guez, Laurent Sifre,
  George Driessche, Julian Schrittwieser, Ioannis Antonoglou, Veda
  Panneershelvam, Marc Lanctot, Sander Dieleman, Dominik Grewe, John Nham, Nal
  Kalchbrenner, Ilya Sutskever, Timothy Lillicrap, Madeleine Leach, Koray
  Kavukcuoglu, Thore Graepel, and Demis Hassabis.
\newblock Mastering the game of go with deep neural networks and tree search.
\newblock {\em Nature}, 529:484--489, 01 2016.

\bibitem{efficientnet}
Mingxing Tan and Quoc Le.
\newblock Efficientnet: Rethinking model scaling for convolutional neural
  networks.
\newblock In {\em International conference on machine learning}, pages
  6105--6114. PMLR, 2019.

\bibitem{criss}
Chau Tran, Yuqing Tang, Xian Li, and Jiatao Gu.
\newblock Cross-lingual retrieval for iterative self-supervised training.
\newblock {\em Advances in Neural Information Processing Systems},
  33:2207--2219, 2020.

\bibitem{speach}
Zolt{\'a}n T{\"u}ske, George Saon, and Brian Kingsbury.
\newblock On the limit of english conversational speech recognition.
\newblock {\em arXiv preprint arXiv:2105.00982}, 2021.

\bibitem{protein}
Juami H.~M. van Gils, Erik van Dijk, Ali May, Halima Mouhib, Jochem Bijlard,
  Annika Jacobsen, Isabel Houtkamp, K.~Anton Feenstra, and Sanne Abeln.
\newblock Introduction to protein folding, 2023.

\end{thebibliography}
}

\end{document}